\documentclass[runningheads]{llncs}

\usepackage{eccv}

\usepackage{eccvabbrv}

\usepackage{graphicx}
\usepackage{booktabs}
\usepackage{tabularx}

\usepackage{slashbox}

\usepackage[accsupp]{axessibility}  

\usepackage{colortbl}

\definecolor{headercolor}{gray}{0.82} 
\definecolor{lightgray}{gray}{0.93}

\usepackage{hyperref}

\usepackage{orcidlink}

\usepackage[framemethod=TikZ]{mdframed}
\definecolor{midGray}{gray}{0.40}
\definecolor{lightYellow}{RGB}{254,254,237}
\mdfdefinestyle{citationFrame}{%
  linecolor=midGray,
  linewidth=0.5pt,
  roundcorner=3pt,
  skipabove=5pt,
  skipbelow=0pt,
  innertopmargin=5pt,
  innerbottommargin=5pt,
  innerrightmargin=5pt,
  innerleftmargin=5pt,
  leftmargin=0pt,
  rightmargin=0pt,
  backgroundcolor=gray!10
}

\begin{document}

\setlength{\abovedisplayskip}{4pt}
\setlength{\belowdisplayskip}{4pt}

\title{Synthetic to Authentic:\\Transferring Realism to 3D Face Renderings for Boosting Face Recognition}

\vspace{-10pt}

\titlerunning{Synthetic to Authentic}

\author{Parsa Rahimi\inst{1,2}\orcidlink{0000-0001-7927-268X} \and Behrooz Razeghi\inst{2}\orcidlink{0000-0001-9568-4166} \and
S\'{e}bastien~Marcel\inst{2,3}\orcidlink{0000-0002-2497-9140}}

\authorrunning{P.~Rahimi \textit{et al.}}

\institute{École Polytechnique Fédérale de Lausanne (EPFL), Lausanne, Switzerland \and
Idiap Research Institute, Martigny, Switzerland \and
Université de Lausanne (UNIL), Lausanne, Switzerland\\
\email{parsa.rahiminoshanahg@epfl.ch,\{behrooz.razeghi,marcel\}@idiap.ch}}

\maketitle

\vspace{-15pt}
\begin{figure}[ht]
    \centering
    \begin{minipage}[b]{0.32\textwidth}
        \includegraphics[width=\textwidth]{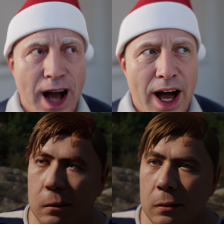}
        \label{fig:first}
    \end{minipage}
    \begin{minipage}[b]{0.32\textwidth}
        \includegraphics[width=\textwidth]{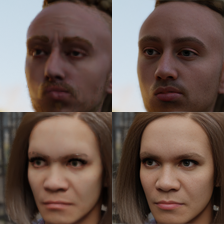}
        \label{fig:second}
    \end{minipage}
    \begin{minipage}[b]{0.32\textwidth}
        \includegraphics[width=\textwidth]{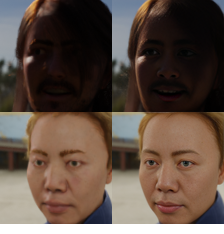}
        \label{fig:second}
    \end{minipage}
    \vspace{-12pt}
        \caption{3D-Rendered images of human faces \cite{bae2023digiface} (left image in each column), and post-processing images by image-to-image translation (right image in each column) for boosting the performance of a Face Recognition trained on the synthetic data.}
    \vspace{-20pt}
\end{figure}


\vspace{-10pt}

\begin{abstract}
In this paper, we investigate the potential of image-to-image translation (I2I) techniques for transferring realism to 3D-rendered facial images in the context of Face Recognition (FR) systems. The primary motivation for using 3D-rendered facial images lies in their ability to circumvent the challenges associated with collecting large real face datasets for training FR systems. These images are generated entirely by 3D rendering engines, facilitating the generation of synthetic identities. However, it has been observed that FR systems trained on such synthetic datasets underperform when compared to those trained on real datasets, on various FR benchmarks. In this work, we demonstrate that by transferring the realism to 3D-rendered images (i.e., making the 3D-rendered images look more real), we can boost the performance of FR systems trained on these more photorealistic images. This improvement is evident when these systems are evaluated against FR benchmarks like IJB-C, LFW which utilize real-world data by $2\%$ to $\%5$, thereby paving new pathways for employing synthetic data in real-world applications. The project page is
available at: \url{https://idiap.ch/paper/syn2auth}.

\end{abstract}

\section{Introduction}
\label{sec:intro}

Given the increasing dependency on artificial intelligence (AI) systems in our everyday lives, it
becomes essential to comprehend and rectify any potential problems that might arise within these
systems. A primary issue with today's systems is their strong dependency on large volumes of data required for training. 
This dependency presents numerous problems, both ethically and legally, in areas such as vision and language models. 
For instance, datasets often collected from web crawls may contain ethically and legally sensitive content, with inherently uncontrollable and inaccurate labels. 
This issue becomes even more critical in the sensitive task of Face Recognition (FR) systems, which requires the collection of personal and sensitive image modalities containing faces.
Furthermore, considering legal policies such as GDPR and other digital ethics guidelines \cite{eugdrppolicy, dicagealg}, the use of existing datasets like WebFace260M \cite{zhu2021webface260m} and CASIA-WebFace \cite{casiawebface} could be problematic when deployed in critical applications.
Besides these concerns, the necessity to collect large sample sizes for training an effective deep FR model poses another challenge. 
Therefore it is crucial to address these issues, which involve one of the most important applications of AI systems in our daily lives: FR systems \cite{kim2022adaface, deng2019arcface} (e.g., unlocking our phones, security gates). 
Due to the mentioned problems with data captured from the real world, 
there has been an increase in research exploring the applicability of synthetic data as an alternative or complement to real datasets in various computer vision problems \cite{li2022bigdatasetgan_syndata_imseg, greff2022kubric_datagen, bae2023digiface}.
For instance, 
studies using a 3D rendering pipeline \cite{wood2021fakeittillyoumakeit} have shown that for tasks like Face Parsing and Landmark Localization, the accurate labels provided by rendering pipelines can surpass the performance of models trained on real datasets in landmark localization tasks (since the images are rendered using a model-based face, the landmark locations are accurate compared to those in human-annotated datasets collected from the real world).

Recently, authors in \cite{azizi2023synthetic} have demonstrated that by using the conditional generation of different classes with a pre-trained denoising diffusion model \cite{sohl2015deep_diff, anderson1982reverse_diff}, it is possible to boost the performance of downstream classification tasks, emphasizing the potential benefits of using synthetic data to enhance AI models.

As mentioned earlier, collecting large datasets for specific computer vision tasks can be challenging, especially in the domain of facial images, which are considered one of the most sensitive data modalities.
To alleviate this problem, there has been a surge in research within the community focused on developing methodologies for creating datasets that either complement existing ones \cite{rahimi2023toward} 
(mainly for bias mitigation, addressing the problem of underrepresented data for some sensitive groups) or 
entirely replace the datasets used for training FR systems. Methods such as IDiffFace \cite{boutros2023idiff}, Digiface1M \cite{bae2023digiface}, and DCFace \cite{kim2023dcface} aim to generate useful datasets for training an FR system from scratch.
To generate a useful dataset for training an FR system, we need to include various identities with diverse demographic labels (i.e., inter-class variability on the order of tens of thousands), and for each identity, variations of the same identity (i.e., intra-class variability, such as different poses and expressions, etc.).
When generating variations of the same identity, it is crucial to ensure the preservation of the identity, for example, when changing the pose. Current methods in literature enforce this condition by using a separate, strong, pre-trained FR system \cite{boutros2023idiff, kim2023dcface} or by utilizing identity attribute labels in large datasets like CASIA-WebFace \cite{casiawebface}. However, the challenge lies in \textit{replacing} the training dataset of the FR system with synthetic data, not using a strong pre-trained FR system trained on real data to generate synthetic data which is a strong and unreasonable prior.

It is difficult to quantify the benefits gained through synthetic datasets, as they often fall short of the performance achievesd by the pre-trained FR systems used during their generation phase.

Another approach involves using 3D-rendering engines, as seen in publicly available datasets like Digiface1M \cite{bae2023digiface}. 
This method is advantageous because it does not require any specific enforcement for identity preservation when generating variations of the same identity, given direct access to the exact mesh and vertices that will eventually be rendered into a face image using different rendering methodologies.

Hence, we can conclude that by changing the pose of the subject or the lighting of the environment, the identity remains unchanged. However, a significant downside is observed when training an FR system with these 3D-rendered datasets, like Digiface1M, and evaluating it against standard FR benchmarks such as IJB-C \cite{ijbc}. There exists a large performance gap, possibly due to an Out-of-Distribution (OOD) problem \cite{bae2023digiface}.

\subsection{Research Problem}

The collection of datasets containing identity-labeled human faces is often impeded by privacy concerns \cite{razeghi2024deepPFmodel}.
Consequently, there is an increasing trend toward synthesizing such data, which is then utilized to train FR models. This paper investigates the following \emph{hypothesis}:

\begin{mdframed}[style=citationFrame,userdefinedwidth=
\linewidth,align=center,skipabove=7pt,skipbelow=0pt]

Face images in existing rendered datasets can be made more realistic while preserving identities, \textbf{without} the need for \textbf{identity labels} or a \textbf{pre-trained FR model}, thereby improving the accuracy of FR models trained on this data.
\end{mdframed}

\noindent
Our primary contribution is to validate this hypothesis through extensive experiments.
It tries to address the OOD problem of 3D Face Renderings compared to face images captured from the real world.

\subsection{Key Contributions}

\begin{figure}
    \centering 
    \includegraphics[width=0.85\textwidth]{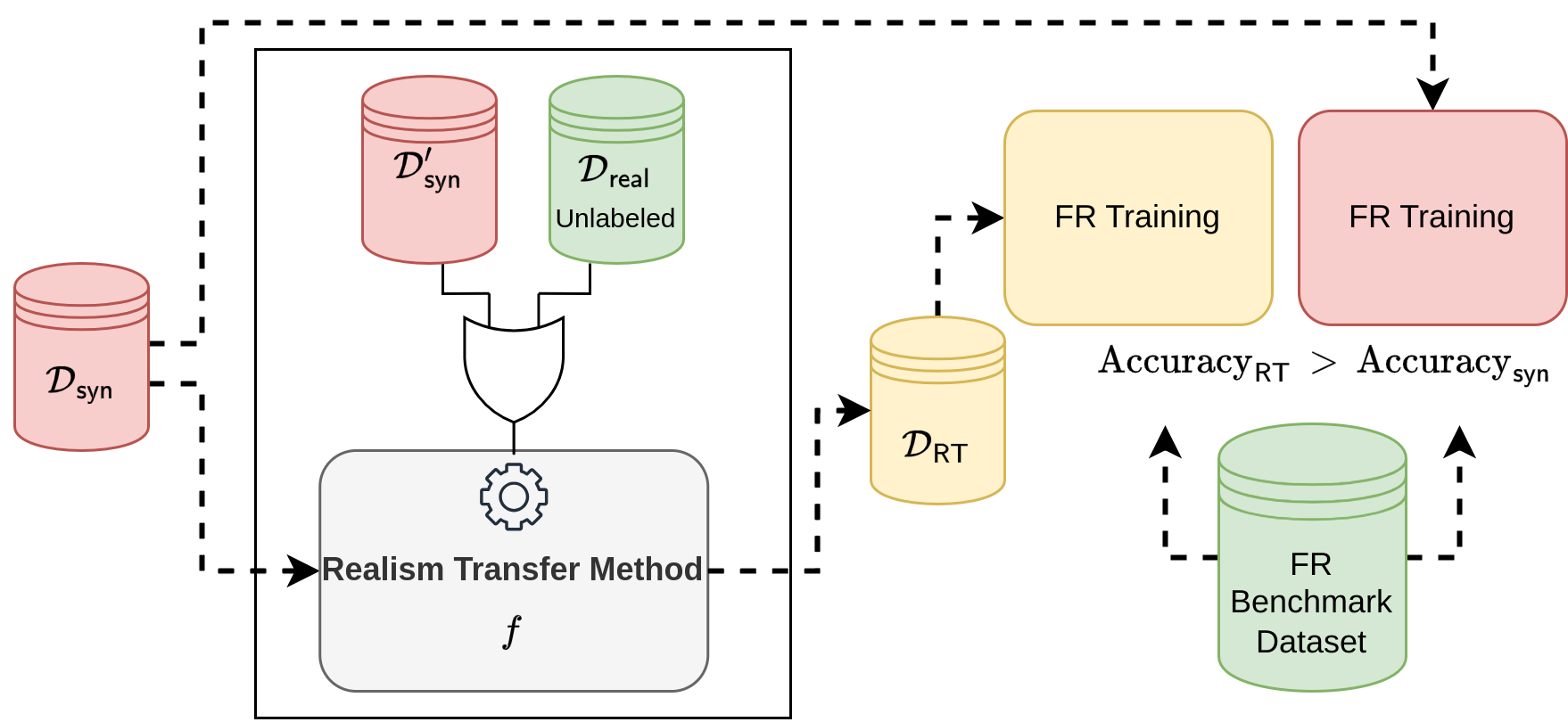}
    \caption{In this paper, we study the efficacy of image-to-image translation methodologies applied to enhance the performance of face recognition---in essence a challenging classification task. 
    Starting with a dataset of 3D-rendered human faces (\emph{i.e.} $\mathcal{D}_{\mathsf{syn}}$),  that exhibit a domain shift compared to real-world human face images, we apply various image-to-image translation and Face Restoration methodologies (\emph{i.e.}, \textbf{Realism Transfer Method} Block) that only require limited 
    \textbf{identity unlabeled} real datasets (\emph{i.e.}, $\mathcal{D}_{\mathsf{real}}$) \textbf{or} subset of unrealistic images (\emph{i.e.}, $\mathcal{D}_{\mathsf{syn}}^{\prime}$) themselves to train. We then train a face recognition network on both the original (unrealistic-looking) and the newly translated (more realistic) images, $\mathcal{D}_{\mathsf{RT}}$, to investigate whether this approach can improve the accuracy of FR systems.}
    \label{fig:overview}
    \vspace{-15pt}
\end{figure}

In this paper, our key contribution lies in investigating and analyzing the potential of introducing photorealism into 3D-rendered datasets, as depicted in \autoref{fig:overview}, \textit{without using any identity labels or a trained FR system}. We demonstrate that we can achieve a performance gain with an FR system trained on our more photorealistic dataset (i.e., transferring realism), thereby opening new avenues for exploring this topic.
To the best of our knowledge, this is the first attempt to study the effect of photorealism on top of 3D-rendered facial images for gaining performance improvement in FR systems. 

Our contributions are as follows:
\vspace{-6pt}
\begin{itemize}
    \item 

    We analyze the applicability of transfer learning methodologies to bridge the gap between imperfect simulation of the real world in the 3D rendering engines, specifically in the domain of face images.
    \item 
    In contrast to previous works, which require a strong pre-trained FR model to generate useful data for training an FR model, we observe a performance boost without relying on any pre-trained FR system or identity labels in the challenging task of FR.

    \item 
    We introduce a mathematical formulation for the realism transfer idea and reformulate other approaches using this unified framework.
\end{itemize}
\vspace{-4pt}
In \autoref{sec:related}, we lay some background on the problem and introduce relevant methods to our analysis. In \autoref{sec:problem}, we define our problem setting. Finally, in \autoref{sec:experiments}, we explain our experimental analysis. 

\section{Related Work} \label{sec:related}

\vspace{-3pt}
In this section, we will briefly overview relevant topics related to the usage and generation of synthetic data, as well as methods that can be applied to our problem setting.

\vspace{-3pt}

\subsection{Synthetic Data in Computer Vision} 

\vspace{-3pt}

Synthetic data generation has become a key strategy for creating vast quantities of accurately annotated data, which is necessary for computer vision tasks that often require detailed labeling. This approach facilitates the development of comprehensive datasets essential for training and improving vision-based algorithms and has been extensively explored in the research community recently. 
For example, synthetic data has been utilized in tasks such as Semantic Image Segmentation \cite{ baranchuk2021label_syndata_imseg, tritrong2021repurposing_syndata_imseg, li2022bigdatasetgan_syndata_imseg}, Optical Flow Estimation \cite{sun2021autoflow_syndata_optflow}, Face Parsing \cite{wood2021fakeittillyoumakeit} and Face Recognition \cite{bae2023digiface}, Human Motion Understanding \cite{guo2022learning_syndata_human, ma2022pretrained_syndata_human}, and other computer vision tasks that require dense, accurate labels. Some of these approaches \cite{wood2021fakeittillyoumakeit, bae2023digiface, greff2022kubric_datagen} utilize 3D-rendering engines and physics simulators \cite{coumans2021_physim} to model the underlying physics of the real world. 
This ensures that the distribution of the generated data is similar to that of data gathered from the real world, making it useful for the applicability of these data and the models trained on them. 
Our analysis in this paper makes a significant stride in alleviating the domain gap caused by imperfect simulation, modeling, and the limited computing power available to simulate the real world.

\vspace{-2pt}

\subsection{Unpaired Image-to-Image Translation}

\vspace{-2pt}

In this section, we briefly highlight methodologies that are particularly promising for enhancing realism in computer graphics applications---a critical challenge in the domain of FR. Among these,
VSAIT \cite{vsait_theiss2022unpaired} introduces a novel method for unpaired image-to-image translation using Vector Symbolic Architectures (VSA) to minimize semantic flipping, which occurs when the content of the translated images does not match the semantic context of source domain. This is specifically important as it plays a key role in the photorealism of computer graphics applications \cite{richter2022enhancing_vsait}. The authors propose leveraging the VSA framework's capacity for high-dimensional symbolic computation to maintain content consistency between the source and translated images. This is especially useful since the VSA framework is robust against noise. This method is one of the methods that we examine for the \emph{Realism Transfer Method} in \autoref{fig:overview}.
In the Density Changing Regularized Unpaired Image Translation (DECENT) method \cite{xie2022unsupervised_im2im},  the authors focused on the concept of density-changing regularization. The method assumes that image patches of high probability density in one domain should be mapped to patches of high density in another domain. To enforce this principle, two density estimators were trained for each domain, and penalties were applied to the variance in density changes. This approach allows for more accurate preservation of neighboring information without relying on pairwise distances.
Recently, authors in \cite{kim2023unsb} introduced the Unpaired Neural Schrödinger Bridge (UNSB) method, which formulates the Schrödinger Bridge problem for the I2I task as a sequence of adversarial learning tasks. By leveraging discriminators and regularization techniques, they effectively overcome the curse of dimensionality. Essentially, their approach minimizes transport costs under constraints of Kullback-Leibler divergence.

\subsection{Inverse Problem and Generative Prior}

\vspace{-4pt}

Among the approaches that incorporate a generative prior, inverse problem methodologies can also be applied to enhance realism. 
We consider two main types of generators: GAN-based and Diffusion-based. Specifically, in the case of employing StyleGANs \cite{sg2,sg3}, which are trained on the domain of real data (\emph{e.g.}, FFHQ \cite{sg1_stylebased_generator} or its recent extension LPFF \cite{wu2023lpff}), and inverting unrealistic images to one of the StyleGANs's latent-spaces (\emph{e.g.}, $\mathcal{W}$, $\mathcal{W}^{+}$) using various methodologies, 
we aim to achieve the desired realism by reconstructing the resulting latent point.
There are various methods for StyleGAN inversion, including \emph{Optimization-Based} \cite{abdal2019image2stylegan, abdal2020image2stylegan++},  \emph{Encoder-Based}, such as e4e~\cite{e4e} and pSp~\cite{psp} and \emph{HyperNetwork-Based} \cite{dinh_hyperinverter_2022}. We leave the interested reader to recent surveys for details of each approach \cite{xia2022gan_inversion}.

Diffusion models \cite{anderson1982reverse_diff, song2020denoising} have recently emerged as a powerful new approach to generative modeling.  
In the diffusion process, these models introduce small amounts of noise to the original image in steps. During the reverse process, they attempt to estimate and remove the noise added to the original image. By repeating this process in the forward phase, we can transition from a signal domain to white Gaussian noise. In the denoising reverse process, it is possible to reconstruct the original signal.
In the context of diffusion models, DDIM inversion \cite{song2020denoising} is a fundamental technique that introduces small increments of noise to a given image to approximate the corresponding input noise. 
Running a reverse diffusion with DDIM and this noise allows for the reproduction of the original image.
In our problem setting, similar to StyleGAN Inversion, we utilize an unconditional diffusion model trained exclusively on a dataset like FFHQ. Our objective is to invert synthetic images back to a noise map and then reconstruct the input image. This approach allows us to uniquely bridge the distribution gap between real and synthetic images.

\vspace{-6pt}
\subsection{Face Restoration Methodologies}
\vspace{-3pt}

Face Restoration in computer vision aims to enhance degraded facial images through methods like super-resolution, denoising, and deblurring. Deep learning models, especially Convolutional Neural Networks (CNNs) and Generative Adversarial Networks (GANs), have shown significant advancements in addressing this problem.
The authors in CodeFormer~\cite{zhou2022codeformer} applied the idea of vector quantization \cite{van2017neural_vector_qunatization} to pre-train a quantized autoencoder through self-reconstruction, thereby obtaining a high-quality discrete codebook of face images and the corresponding decoder. The combination of the codebook's prior knowledge and the decoder is then used for face restoration. Based on this codebook prior, a Transformer is employed for the accurate prediction of code combinations from low-quality inputs. Additionally, a controllable feature transformation module is introduced to enable a flexible trade-off between the restoration quality and fidelity of the downgraded face images. 
The authors in PGDiff~\cite{yang2023pgdiff_nips23} introduced the concept of partial guidance, in which the diffusion prior acts as a regularization, and guidance is provided only on the desired properties of
high-quality images. The key to \cite{yang2023pgdiff_nips23} is constructing proper guidance for each task of restoration, inpainting, and masking separately. Methods like PGDiff \cite{yang2023pgdiff_nips23} cannot 
be directly applied to our problem setting, as they require the use of a pre-trained FR system for their restoration guidance.

\subsection{Synthetic Data Generation for Face Recognition}
\vspace{-7pt}

The authors in SYNFace \cite{qiu2021synface} utilize DiscoFaceGAN \cite{deng2020disentangled_disco} to create facial images with detailed control over specific attributes such as identity, pose, expression, and illumination. This addresses the issue of limited variation within synthetic datasets, which impacts the performance of FR systems. By blending features from two synthetic identities to create new ones,  
SYNFace suggests a method to closely mimic real-world data, recommending a combination of synthetic and real images. In \cite{colbois2021use}, the authors invert a dataset containing binary attribute labels of faces into the $\mathcal{W}$ space of a StyleGAN2 generator. They then fit a Support Vector Machine, using the distance to the hyperplane as a measure of the variation's scale. By moving in the direction perpendicular to the hyperplane for each attribute, they generated a small dataset to evaluate an FR system.
As mentioned earlier, DigiFace-1M \cite{bae2023digiface} provides a large-scale synthetic dataset for FR, produced through computer graphics. It uniquely defines each identity with specific facial details, allowing for varied expressions and environments. This model, which is independent of real data, narrows the gap between synthetic and real data, setting a new benchmark for accuracy. However, it faces challenges such as unrealistic textures and an unexamined demographic distribution. 

DCFace \cite{kim2023dcface}, a newer Diffusion model, is designed for synthetic FR and features a two-stage process: generating synthetic identities and mixing these identities with styles from a ``style bank.'' This approach demonstrates a strong capacity for creating unique and diverse identities, as evidenced by its performance in comparison with other approaches. However, as previously mentioned, the use of pre-trained FR systems or datasets with large identity 
is an unreasonable prior as the goal is to generate synthetic data for training FR system primarily. 
In IDiffFace \cite{boutros2023idiff}, the authors introduced a method for generating synthetic datasets for face recognition by leveraging conditional Latent Diffusion Models (LDM) \cite{rombach2022high_ldm}. Significant emphasis is placed on the diffusion model's conditioning mechanism on face embeddings from a pre-trained FR system. This approach enables the creation of highly realistic and varied synthetic faces by conditioning the generative process on compact, identity-specific embeddings, albeit at the cost of utilizing a separate pre-trained FR system and the identity labels provided by large FR datasets.
GANDiffFace \cite{melzi2023gandiffface} relies on the popular pre-trained model provided by Stable Diffusion. This approach comprises two steps: the first is dedicated to the synthesis of identities based on StyleGAN3 \cite{sg3} and transformation in its latent space. This transformation is based on directions in the latent space that change specific attributes of images to introduce small intra-class variability, such as altering the pose. Subsequently, relying on the pre-trained text-to-image generator Stable Diffusion and the DreamBooth \cite{ruiz2023dreambooth} personalization fine-tuning approach, they introduce more intra-class variability. The problem with this approach is its high reliance on large datasets \cite{schuhmann2022laion5b} used to train Stable Diffusion, which are not privacy-friendly.

\vspace{-9pt}

\section{Transferring Realism to 3D Rendered Faces} \label{sec:problem}

\vspace{-7pt}
\subsection{Problem Formulation}
\vspace{-3pt}

Consider a dataset $\mathcal{D}_{\mathsf{syn}}$ comprising 3D-rendered images $\{ \{ \mathbf{x}_n^{k} \}_{k=1}^{K_n} \}_{n=1}^{N} \subseteq \mathcal{X}$ of human faces, consisting of $N$ identities. For each identity, $n \in \{1, \cdots, N\}$, there exists an identity-dependent number of variations, $K_n$, representing different variations of the same identity. Let $P_{\mathbf{X}}$ denote the empirical probability distribution of the synthetic 3D-rendered data.

Our objective is to improve the utility of the synthetic dataset $\mathcal{D}_{\mathsf{syn}}$ with respect to a utility measure, by utilizing either an unlabeled real dataset $\mathcal{D}_{\mathsf{real}}$ with few samples, or a subset of the synthetic dataset $\mathcal{D}_{\mathsf{syn}}$ itself, denotes as $\mathcal{D}^{'}_{\mathsf{syn}}$, for training  FR systems. 
In the following, we explore various approaches to post-process the synthetic dataset $\mathcal{D}_{\mathsf{syn}}$ to obtain a new dataset $\mathcal{D}_{\mathsf{RT}}$ for training the FR systems. We denote the generating distribution of post-processed data by $P_{\mathbf{Y}}$.



Consider two measurable spaces $\mathcal{X}$ and $\mathcal{Y}$, where $\mathcal{X}$ represents the domain of 3D-rendered images (source), and $\mathcal{Y}$ represents the domain of images captured from the real world (target). Let $\mathbf{X} \sim P_{\mathbf{X}}$ and $\mathbf{Y} \sim P_{\mathbf{Y}}$ be random objects representing random realizations from these spaces, with distributions $P_{\mathbf{X}}$ and $P_{\mathbf{Y}}$ respectively, where $\mathbf{X} \in \mathcal{X}$ and $\mathbf{Y} \in \mathcal{Y}$. 
Let $f : \mathcal{X} \rightarrow \mathcal{Y}$ denote a mapping function that transforms elements from the source domain to the target domain, and let $g : \mathcal{Y} \rightarrow \mathcal{X}$ denote a mapping function for the reverse transformation. These mappings can be implemented as deep neural networks due to their flexibility and capacity for learning complex transformations. However, our study primarily focuses on the forward mapping $f : \mathcal{X} \rightarrow \mathcal{Y}$, which transforms elements from the source domain $\mathcal{X}$ to the target domain $\mathcal{Y}$.

The objective of the image-to-image translation problem is to learn (find) these mappings $f$ and $g$ such that: \textbf{(i)} the distribution of the mapped object approximates the distribution of the target object, i.e., $P_{f \left(\mathbf{X}\right)} \approx P_{\mathbf{Y}}$ and/or $P_{\mathbf{X}} \approx P_{g\left(\mathbf{Y}\right)}$; and \textbf{(ii)} the mapping preserves or captures specific characteristics or features of the input images. This objective can be formally expressed as a constraint optimization problem, where the mapped images maintain certain predefined properties or metrics of similarity with the input images, fundamental to tasks like style transfer, domain adaptation, or generative modeling.

Let $\mathsf{dist} \left( P_{f\left(\mathbf{X}\right)} , P_{\mathbf{Y}} \right)$ denote a discrepancy measure between the distributions of the transformed source images and the target images. For example, one can use the $\mathsf{f}$-divergence $\mathsf{dist} ( P_{f\left(\mathbf{X}\right)}, P_{\mathbf{Y}} ) = \mathrm{D}_{\mathsf{f}} (P_{f\left(\mathbf{X}\right)} \Vert P_{\mathbf{Y}} )$ as such a measure. The optimization problem then aims to minimize a loss function that quantifies both the distributional similarity and the preservation of image characteristics:
\begin{equation}\label{I2I_GeneralizedProblemFormulation}
    \mathop{\min}_{f, g}\;\;\;
    \mathsf{dist} \left( P_{f\left(\mathbf{X}\right)} , P_{\mathbf{Y}} \right) + 
    \mathsf{dist} ( P_{g\left(\mathbf{Y}\right)} , P_{\mathbf{X}} )+ 
    \lambda_\mathsf{x} \Phi_\mathsf{x} ( \mathbf{X}, f \left(\mathbf{X}\right) ) +
    \lambda_\mathsf{y} \Phi_\mathsf{y} ( \mathbf{Y}, g \left(\mathbf{Y}\right) ),   
\end{equation}
where $\Phi_\mathsf{x}$ and $\Phi_\mathsf{y}$ are penalty functions that enforce the preservation of desired features in the transformed images, with $\lambda_\mathsf{x}$ and $\lambda_\mathsf{y}$ balancing the importance of distribution similarity and feature preservation.

\subsection{Applying General Formulation to Related Works}

\vspace{-3pt}

\noindent
\textbf{DECENT \cite{xie2022unsupervised_im2im}:}
The DECENT objective is introduced as:
\begin{equation}
\mathop{\min}_{f}\;\;\;
\mathcal{L}_{\mathsf{gan}} + \lambda_{\mathsf{identity}} \, \mathcal{L}_{\mathsf{identity}} +
\lambda_{\mathsf{density}} \, \mathcal{L}_{\mathsf{density}},
\end{equation}
where $\mathcal{L}_{\mathrm{gan}} \! =
\mathbb{E}_{P_{\mathbf{X}}} \left[ \log ( 1 \! - \! D \! \left(f \! \left( \mathbf{X}\right) \right)  \right] + \mathbb{E}_{P_{\mathbf{Y}}} \! \left[  \log D \! \left( \mathbf{Y} \right) \right]$, $\mathcal{L}_{\mathsf{identity}} \! = \mathbb{E}_{P_{\mathbf{Y}}} \! \left[ f(\mathbf{Y}) \!- \! \mathbf{Y} \right]$, and $\mathcal{L}_{\mathsf{density}} = \mathsf{V} \left( \frac{h_{\mathcal{X}} \left( \mathbf{X} \right) }{ h_{\mathcal{Y}} \left( f \left( \mathbf{X} \right) \right)}\right)$, with $\mathsf{V}$ as the variance function, 
$h_{\mathcal{X}}$ and $h_{\mathcal{Y}}$ being density estimators for the corresponding domains, and $D$ as the discriminator (scoring function).

Given our general problem formulation as described in equation \eqref{I2I_GeneralizedProblemFormulation}, it's important to note that in many I2I translation models---particularly those influenced by the CycleGAN framework---the functions $f$ and $g$ work together to enforce cycle consistency.
This means that for any image $\mathbf{X} \in \mathcal{X}$, the transformation sequence $\mathbf{X} \rightarrow f(\mathbf{X}) \rightarrow g(f(\mathbf{X}))$ should closely approximate $\mathbf{X}$. Similarly, this principle applies in reverse, ensuring that the mappings $f$ and $g$ function as approximate inverses of one another. This preserves the content of the images while facilitating translation between domains. Therefore, the identity loss $\mathcal{L}_{\mathsf{identity}}$ is strategically implemented to reinforce this principle by encouraging the function $f$ to act as an identity map when provided with inputs from its target domain $\mathcal{Y}$. 
Thus, $\mathcal{L}_{\mathrm{gan}}$ corresponds to $\mathsf{dist} \left( P_{f\left(\mathbf{X}\right)} , P_{\mathbf{Y}} \right) $, $\mathcal{L}_{\mathsf{identity}} $ corresponds to $ \Phi_\mathsf{y} ( \mathbf{Y}, g \left(\mathbf{Y}\right) )$, and $\mathcal{L}_{\mathsf{density}}$ corresponds to  $\Phi_\mathsf{x} ( \mathbf{X}, f \left(\mathbf{X}\right) )$ in \eqref{I2I_GeneralizedProblemFormulation}.

\vspace{2pt}
\noindent
\textbf{VSAIT \cite{vsait_theiss2022unpaired}:}
The VSAIT objective is introduced as follows:
\begin{equation}
\mathop{\min}_{f}\;\;\;
\mathcal{L}_{\mathsf{gan}} + \lambda \, \mathcal{L}_{\mathsf{VSA}},
\end{equation}
where $\mathcal{L}_{\mathsf{gan}}$ represents the hypervector adversarial loss, aimed at aligning the distribution of generated images with that of the target images. Meanwhile, $\mathcal{L}_{\mathsf{VSA}}$
is a loss designed to ensure the generator preserves the source content and minimizes semantic flipping.

\vspace{2pt}
\noindent
\textbf{UNSB \cite{kim2023unsb}:}
In the context of our general problem formulation, the Schrödinger Bridge for image-to-image translation is tailored to find a mapping $f: \mathcal{X} \rightarrow \mathcal{Y}$ that minimizes:\vspace{-2pt}
\begin{equation}
\mathop{\min}_{f}\; \mathsf{dist} \left( P_{f\left(\mathbf{X}\right)} , P_{\mathbf{Y}} \right) + \lambda \; \Phi_\mathsf{x} (\mathbf{X}, f(\mathbf{X})),
\end{equation}
where $\mathsf{dist} \left( P_{f\left(\mathbf{X}\right)} , P_{\mathbf{Y}} \right) = \mathrm{D}_{\mathsf{KL}} (P_{f(\mathbf{X})} \| P_{\mathbf{Y}})$ is the Kullback-Leibler divergence.

\vspace{2pt}
\noindent
\textbf{CodeFormer \cite{zhou2022codeformer}:}
The objective of CodeFormer is introduced as follows:
\begin{equation}
\mathop{\min}_{f}\;\;\;
\mathcal{L}_{\mathsf{L_1}} + \mathcal{L}_{\mathsf{perceptual}} + \mathcal{L}_{\mathsf{code}}  + \lambda_{\mathsf{gan}} \, \mathcal{L}_{\mathsf{gan}},
\end{equation}
where $\mathcal{L}_{\mathsf{L_1}}$ represents the $L_1$ loss in the image domain (between source and targeted images), $\mathcal{L}_{\mathsf{perceptual}}$ denotes the $L_2$ loss in the embedding space (between embeddings of the source and target images), $\mathcal{L}_{\mathsf{code}}$ is the $L_2$ loss of codeword approximations, and $\mathcal{L}_{\mathsf{gan}}$ is the typical adversarial loss between the source image and the reconstructed image. Considering our general problem formulation \eqref{I2I_GeneralizedProblemFormulation}, the $\mathcal{L}_{\mathsf{L_1}}$ and $\mathcal{L}_{\mathsf{gan}}$ losses contribute towards the $\mathsf{dist} \left( P_{f\left(\mathbf{X}\right)} , P_{\mathbf{Y}} \right) + \mathsf{dist} ( P_{g\left(\mathbf{Y}\right)} , P_{\mathbf{X}} )$ terms, while the other terms act as penalty functions. For more details, we refer the readers to Section~6 of \cite{razeghi2023bottlenecks}, where the authors address generative compression techniques from the perspective of the transform coding problem and the classical Shannon rate-distortion theorem.

\vspace{2pt}
\noindent
\textbf{DDIM Inversion \cite{song2020denoising}:}
The objective of DDIM Inversion can be aligned with the general problem formulation in equation \eqref{I2I_GeneralizedProblemFormulation} by introducing an optimization problem that seeks to minimize:\vspace{-1pt}
\begin{equation}\label{DDIM_inversion}
    \mathop{\min}_{f}\;\;\;
    \mathsf{dist} \left( P_{f\left(\mathbf{X}\right)} , P_{\mathbf{Y}} \right) + 
    \lambda_\mathsf{x} \Phi_\mathsf{x} ( \mathbf{X}, f \left(\mathbf{X}\right) ).
\end{equation}

Having outlined the brief theoretical underpinnings and methodological frameworks for enhancing the realism of synthetic 3D imagery, we now proceed to empirically validate these approaches through a series of experiments designed to assess their efficacy in practical applications.


\vspace{-10pt}

\section{Experiments}
\label{sec:experiments} 

\vspace{-7pt}

\subsubsection*{Methodology:} For transferring realism, we began by exploring various methods mentioned in \autoref{sec:related}, namely, CodeFormer\cite{zhou2022codeformer}, VSAIT \cite{vsait_theiss2022unpaired}, UNSB\cite{kim2023unsb}, Decent\cite{xie2022unsupervised_im2im}, DDIM Inversion and StyleGAN Inversion \cite{xia2022gan_inversion}. The goal is to apply Realism Transfer methods to unrealistic images (\emph{i.e.}, $\mathcal{D}_{\mathsf{syn}}$) to get more photorealistic versions (\emph{i.e.}, $\mathcal{D}_{\mathsf{RT}}$). These versions are used for training and evaluating a FR system. For evaluation, we report verification accuracy (\emph{i.e.}, True Acceptance Rate (TAR)), where the thresholds are set using cross-validation \cite{huang2008labeled_lfw_easy} (see \autoref{Table:PerforamnceMetrics_Easy}), and TARs at different thresholds determined by fixed False Match Rates (FMR) in \autoref{Table:PerformanceMetrics_FR_IJBC} \cite{ijbc}.  

\vspace{-12pt}
\subsubsection*{Experiment Setup:} In the case of the CodeFormer, 
we utilized the pre-trained models provided by the authors. These models were trained solely on the FFHQ dataset \cite{sg1_stylebased_generator}, which does not contain any identity labels, as the identity information was not used in their restoration method.

\begin{table*}[b!]
    \vspace{-7pt}
    \caption{Processing time (\emph{i.e.}, Time/image(s)) and Qualitative Image quality assessment (\emph{i.e.}, Qualitative Ex) of different realism transfer methods.}\label{table:final_pick}
    \vspace{-7pt}
    \centering
    \scalebox{0.6}{
    \begin{tabular}{|c|c|c|c|c|c|}
    \hline
    
    \backslashbox{Metric}{ Method} & \cellcolor{lightgray} CodeFormer & \cellcolor{lightgray} VSAIT & \cellcolor{lightgray} DECENT & \cellcolor{lightgray} UNSB & \cellcolor{lightgray} DDIM Inversion \\
    \hline \hline
    \cellcolor{lightgray}Time/Image (s) & 0.41 & 0.015 & 0.13 & 0.38 & 8.7 \\
    \hline
    \cellcolor{lightgray} Qualitative Ex & Good & Average & Average & Average & Good \\
    \hline
    \end{tabular}}
\end{table*}

\begin{figure}
    \centering
    \includegraphics[width=\textwidth]{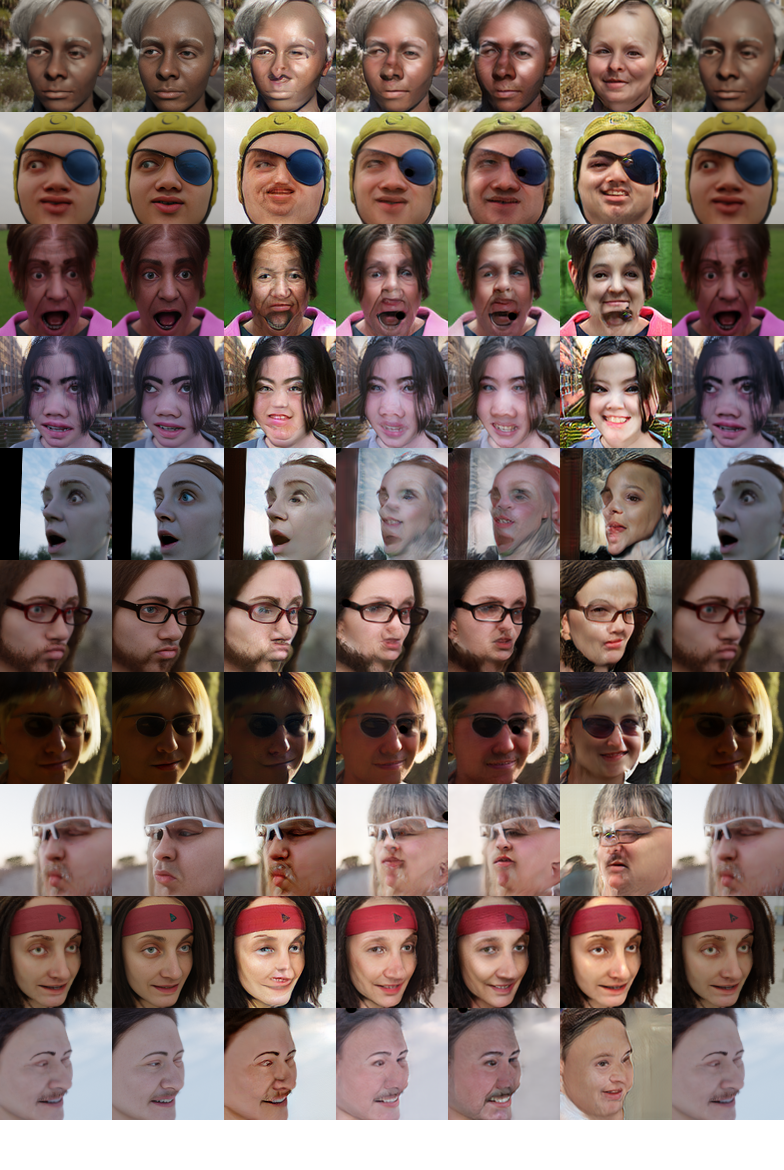}
    \vspace{-30pt}
    \caption{From the left to right, the first column corresponds to the original DigiFace1M dataset, and the next columns are from after applying different translation tasks to the original images, CodeFormer, VSAIT, DECENT, UNSB-NE-1, UNSB-NE-5 and DDIM Inversion, respectively.}
    \label{fig:side_by_side}
\end{figure}

For training unpaired I2I methods, specifically, VSAIT, UNSB, and DECENT, we randomly selected five shards for the source domain (\emph{i.e.}, 3D-rendered human face images), each containing $20,000$ images from the DigiFace1M dataset. Similarly, for the target domain, we randomly selected five shards, each containing $20,000$ images from the FFHQ dataset, and experimented with training these three models using multiple combinations of source and target shards.
After training the realism transfer methods, we selected two according to the time they needed to process an image and qualitative examination of the processed images, which are depicted by \emph{Time/Image (s)} and \emph{Qualitative Ex} respectively in \autoref{table:final_pick}. The processing time was measured on an NVIDIA RTX 3090 Ti across all methods. 
\autoref{fig:side_by_side} presents some qualitative results of various methods. As can be qualitatively observed from \autoref{fig:side_by_side}, CodeFormer generally performed very well across all samples, preserving the entire facial structure. In contrast, VSAIT, DECENT, and UNSB did not consistently produce quality images. 
Notably, these models sometimes dislocated parts of the images, resulting in multiple eyes and mouths. Surprisingly, as we will demonstrate in the next section, VSAIT boosted the performance of FR systems. 
Here, `UNSB-NE-1' and `UNSB-NE-5' refer to the number of Neural Evaluation (NE) steps of the method; for more details, please refer to the original paper. Finally, images produced by DDIM-Inversion appear as smoothed-out versions of the originals.
Among the examined methods, we chose CodeFormer because of its good quality and reasonable compute time and VSAIT for its lower compute time and slightly better quality than other I2I methods for the final FR experiments in the next section.

\vspace{-10pt}
\subsection{Face Recognition Experiments}
\vspace{-5pt}

For a fair comparison between different methods, we trained an FR system consisting of a ResNet50 backbone as modified in ArcFace's implementation \cite{deng2019arcface}, with the AdaFace \cite{kim2022adaface} head for contrastive loss. 
We trained a separate network 
for each of the methods mentioned in the previous section, namely, the original DigiFace1M, and translated versions of the images generated using CodeFormer and VSAIT. We name the translated  dataset \textbf{RealDigiFace}. We also included an FR baseline that methods like DCFace and IDiffFace are using; we used the pre-trained model provided by the AdaFace paper, which was trained on the WebFace4M dataset.
For FR benchmarking, we considered various datasets including
LFW \cite{huang2008labeled_lfw_easy},
CFPFP \cite{sengupta2016frontal_cfpfp_easy},
CPLFW \cite{zheng2018cross_cplfw_easy},
CALFW \cite{zheng2017cross_calfw_easy},
AgeDB \cite{moschoglou2017agedb_agedb_easy}, which consist of high-quality images with various lighting, poses, and ages. We also benchmarked against IJB-C \cite{ijbc}, which is amongst the most challenging FR benchmarks in the literature. The results are reported in \autoref{Table:PerforamnceMetrics_Easy} and \autoref{Table:PerformanceMetrics_FR_IJBC}. 

In the tables mentioned, the first column, \emph{Transfer Method}, refers to the translation method used to translate the dataset. For example, if we want to translate the DigiFace1M dataset using CodeFormer, the \emph{Transfer Method} column for the row corresponding to this experiment is set to CodeFormer. For the case of the WebFace4M, IDiffFace, and DCFace, we did not apply the translation, as expected, since they are not 3D-Rendered data, and we wanted to compare with these datasets as is. The \textit{Type} column refers to the nature of the dataset, which can be either \emph{Real} (collected from the real world), \emph{Syn}  (synthetically generated), or \emph{Syn-RT} (translated from a \emph{Syn} dataset using the method mentioned in the \emph{Transfer Method}). 
The \emph{SynGen Req} columns depict whether the \emph{Transfer Method} or the method used for generating the DigiFace1M, DCFace, and IDiffFace dataset requires the identity labels or a pre-trained FR system. 
Here, \emph{No-Req} means that neither the translation method nor the method used to generate the original dataset (i.e., DigiFace1M in our experiments) requires the identity label or pre-trained FR system, and \emph{Pre-Trained FR} indicates that generating the dataset required a pre-trained FR system, which is \emph{undesirable} for the problem setting.

\begin{table*}[bp]
    \vspace{-7pt}
    \caption{Results of different synthetic data generation methodologies used to train multiple FR systems evaluated on the LFW, CFPFP, CPLFW, CALFW, and AgeDB, the last column is the average test accuracy over these five datasets. We are reporting \emph{mean} and \emph{std} over multiple runs of experiments in each row.}
    \vspace{-7pt}
    \label{Table:PerforamnceMetrics_Easy}
    \centering
    \scalebox{0.975}{
    \setlength{\tabcolsep}{0.4em} 
    \renewcommand{\arraystretch}{1.6}
    \resizebox{\linewidth}{!}{%
    \begin{tabular}{|c|c|c|c|c|c|c|c|c|c|}
    \hline
    \rowcolor{lightgray}
    \multicolumn{1}{|c|}{\textbf{Transfer Method}} & \multicolumn{1}{c|}{\textbf{Dataset}} & \multicolumn{1}{c|}{\textbf{Type}} & \multicolumn{1}{c|}{\textbf{SynGen Req}} & \multicolumn{1}{c|}
    {\textbf{LFW}} & \multicolumn{1}{c|}{\textbf{CFPFP}} & \multicolumn{1}{c|}{\textbf{CPLFW}} & \multicolumn{1}{c|}{\textbf{CALFW}} & \multicolumn{1}{c|}{\textbf{AGEDB}} & \multicolumn{1}{c|}{\textbf{Avg}}\\\hline
    \hline
    None & WebFace4M & Real & -  & \textbf{99.78±0.00}	& \textbf{98.97±0.00}	& \textbf{94.17±0.00} & \textbf{95.98±0.00} & \textbf{97.78±0.00} & \textbf{97.34±0.00} \\
    \hline
    \hline
    \rowcolor{lightgray}
    None & DigiFace1M & Syn & No-Req & 91.29±0.57 & 88.62±0.69	& 70.28±0.42 & 73.38±1.15 & 68.24±2.17 & 78.14±0.84 \\\hline
    
    VSAIT \cite{vsait_theiss2022unpaired} & DigiFace1M & Syn-RT & No-Req & 92.87±0.15 & 90.25±0.17 & 72.91±0.68 & 75.98±0.28 & \textbf{\emph{70.83±1.22}} & 80.32±0.25 \\\hline
    
    \rowcolor{lightgray}
    CodeFormer \cite{zhou2022codeformer} & DigiFace1M & Syn-RT & No-Req  & \textbf{\emph{93.07±0.27}} & \textbf{\emph{90.50±0.26}} & \textbf{\emph{73.02±0.62}} & \textbf{\emph{76.59±0.19}}	 & 70.19±2.57 & \textbf{\emph{80.40±0.29}} \\
    \hline
    \hline 
    None & IDiffFace\cite{boutros2023idiff} & Syn & {\color{red}Pre-Trained FR} & 96.37±0.15 & 95.54±0.11 & 73.00±0.47 & 86.24±0.29 & 78.29±0.63 & 84.58±0.16	\\
    \hline
    \rowcolor{lightgray}
    None & DCFace\cite{kim2023dcface} & Syn & {\color{red}Pre-Trained FR}  & 97.94±0.14 & 97.87±0.08 & 78.99±0.49	& 90.35±0.30 & 87.46±0.46	& 89.51±0.13 \\

    \hline
    \end{tabular}}
    }
    \vspace{-5pt}
\end{table*}

\begin{table*}[b!]
    \caption{Results of different synthetic data generation methodologies used to train multiple FR systems evaluated on the IJB-C benchmark, here the numbers in the header of the last six columns represent the different TAR@FMR. We are reporting \emph{mean} and \emph{std} over multiple runs of experiments in each row. }
    \vspace{-7pt}
    \label{Table:PerformanceMetrics_FR_IJBC}
    \centering
    \scalebox{0.975}{
    \setlength{\tabcolsep}{0.4em} 
    \renewcommand{\arraystretch}{1.6}
    \resizebox{\linewidth}{!}{%
    \begin{tabular}{|c|c|c|c|c|c|c|c|c|c|}
    \hline
    \rowcolor{lightgray}
    \multicolumn{1}{|c|}{\textbf{Transfer Method}} & \multicolumn{1}{c|}{\textbf{Dataset}} & \multicolumn{1}{c|}{\textbf{Type}} & \multicolumn{1}{c|}{\textbf{SynGen Req}} & \multicolumn{1}{c|}
    {\textbf{1e-06}} & \multicolumn{1}{c|}{\textbf{1e-05}} & \multicolumn{1}{c|}{\textbf{1e-04}} & \multicolumn{1}{c|}{\textbf{0.001}} & \multicolumn{1}{c|}{\textbf{0.01}} & \multicolumn{1}{c|}{\textbf{0.1}}\\\hline
    \hline
    None & WebFace4M & Real & -  & \textbf{91.78±0.00}	& \textbf{95.22±0.00} & \textbf{96.98±0.00}	& \textbf{98.14±0.00} &	\textbf{98.84±0.00} & \textbf{99.40±0.00}	\\
    \hline
    \hline
    \rowcolor{lightgray}
    None & DigiFace1M & Syn & No-Req  & 18.80±3.83 & 28.96±5.16 & 41.35±5.32 & 56.38±5.18	&72.11±4.30	&87.55±2.76 \\\hline
    VSAIT \cite{vsait_theiss2022unpaired} & DigiFace1M & Syn-RT & No-Req  & 20.14±2.98	& 30.94±2.82 &	43.98±2.88	&59.84±2.12	&75.69±1.67	& 90.21±0.67 \\\hline
    \rowcolor{lightgray}
    CodeFormer \cite{zhou2022codeformer} & DigiFace1M & Syn-RT & No-Req  & \textbf{\emph{23.72±2.76}}& \textbf{\emph{32.48±3.53}} & \textbf{\emph{45.88±3.53}} &	\textbf{\emph{61.74±2.61}}	& \textbf{\emph{77.27±1.54}} & \textbf{\emph{90.72±0.81}} \\
    \hline
    \hline 
    None & IDiffFace\cite{boutros2023idiff} & Syn & {\color{red}Pre-Trained FR}  & 
    29.21±3.97	& 41.77±2.66 & 56.19±1.51 & 71.42±0.79	& 85.59±0.33 & 95.44±0.02	\\
    \hline
    \rowcolor{lightgray}
    None & DCFace\cite{kim2023dcface} & Syn & {\color{red}Pre-Trained FR}  &
    40.89±0.21 & 58.58±1.93	& 72.69±0.53 & 83.80±0.00	& 91.97±0.12 &	97.25±0.02	 \\
    \hline
    \end{tabular}}
    }
\end{table*}

We want to emphasize that we repeated the experiments two to four times, reporting the \emph{mean} and \emph{standard deviation} (std) across all benchmarks (\emph{i.e.}, if we observed high variance we repeated the experiment), and also trained the FR models under the same settings (i.e., all were trained using an IR50 backbone and AdaFace head, with the same early stopping procedure, etc.) for a fair comparison and to ensure that the conclusions drawn are more reliable.

In the case of \autoref{Table:PerforamnceMetrics_Easy}, compared to the model trained on the DigiFace1M, we observed an average improvement of $2.0\%$ over the FR model trained on images generated after transferring them using CodeFormer and VSAIT, with CodeFormer demonstrating a slight advantage in all datasets: LFW, CFPFP, CPLFW, CALFW, and AgeDB. However, it can be observed that there is a significant performance gap between the model trained on the WebFace4M and all other methods, including the DCFace and IDiffFace models, which use such a competitive FR system for generating their images.

\begin{figure}[t!]
    \centering
    \includegraphics[width=0.6\textwidth]{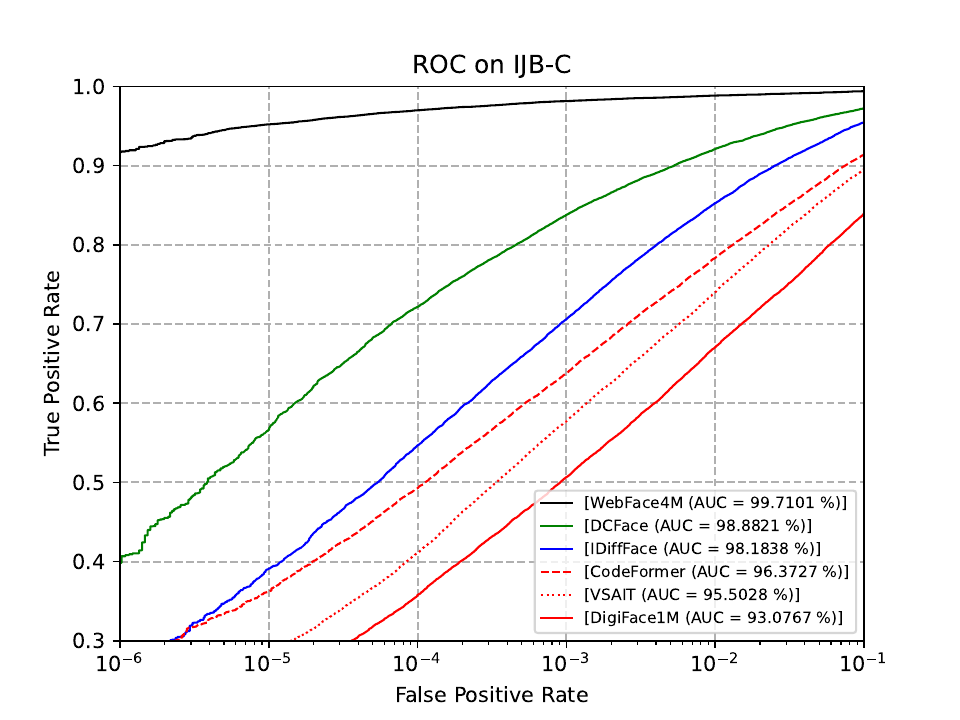}
    \vspace{-6pt}
    \caption{ROC Curve on the IJB-C benchmark, for each dataset we selected one of the models in which we trained an FR on top of it, and plotted the ROC curve.}
    \vspace{-8pt}
    \label{fig:roc_sample}
\end{figure}

In the challenging benchmark of IJB-C, as shown in \autoref{Table:PerformanceMetrics_FR_IJBC}, we first want to 
highlight the significant gap in performance between a strong FR system trained on the WebFace4M dataset across all FMR values, and both DCFace and IDiffFace, as well as our models trained on the transferred images. 
The performance boost observed of Realism images in the IJB-C across different FMR values is larger than that reported in \autoref{Table:PerforamnceMetrics_Easy}, with an average gain of about $                                  3-5\%$. 

The performance boost of Realism Transfer is notably larger at lower FMR values. Specifically, for models trained on images translated using CodeFormer, the performance approaches that of DCFace and IDiffFace at lower FMRs.
Further, we also plotted ROC Curves in the \autoref{fig:roc_sample}, is also emphasizes that models trained on synthetic data are lagging far behind the models that are trained on real data (\emph{i.e.}, WebFace4M), we can also clearly observe the performance boost of Realism Transfer with respect to the DigiFace1M baseline.

\vspace{-5pt}

\vspace{-10pt}
\section{Conclusion and Future Work}

\vspace{-10pt}

In this paper, we have explored the potential of utilizing various I2I and face restoration methodologies to address the challenges posed by imperfect rendering in 3D-rendered FR datasets, with the aim of making them more realistic.

Surprisingly, we found that by solely employing transfer models that do not incorporate identity labels in their training paradigm, we can boost the performance of FR systems across all benchmarks—LFW, CFPFF, CPLFW, CALFW, AGEDB, and IJB-C—by $2\%$ to $5\%$. This improvement is observed in comparison with models trained on the original DigiFace1M dataset, thereby narrowing the performance gap with models that use pre-trained FR data for generating their data. Moreover, this approach moves us closer to our ultimate goal of achieving performance parity with models trained on real data. This opens new avenues for exploring the use of transfer methodologies in the domain of data enhancement for improved downstream model performance.
Given that the pipeline for developing a new transfer method and applying it to the entirety of a source dataset is cumbersome and time-consuming---especially since it necessitates multiple trainings of the FR system on the generated data for conclusions---a future work, could be to explore a form of quality assessment metric. This metric would correlate with the final performance of the FR system when trained on the generated dataset, allowing for the evaluation of the transferred data independently. This approach could significantly streamline the process of assessing the potential of newly generated datasets for FR applications.

\vspace{-10pt}

\subsubsection*{Acknowledgment}

This research is based on work conducted in the SAFER project and supported by the Hasler Foundation's Responsible AI program. 
We would also like to extend our gratitude to Dr.~Damien~Teney for his thoughtful feedback.

%
%


\bibliographystyle{splncs04}
\bibliography{main}
\end{document}